\documentclass[10pt]{article}

\usepackage[letterpaper,margin=1in]{geometry}

\usepackage[T1]{fontenc}
\usepackage{lmodern}

\usepackage{amsmath,amsfonts,bm}

\usepackage{graphicx}
\usepackage{booktabs}

\usepackage{natbib}
\setcitestyle{authoryear,round,citesep={;},aysep={,},yysep={;}}

\usepackage{authblk}

\usepackage{hyperref}
\usepackage{url}

\title{Extrapolation of Periodic Functions Using Binary Encoding of Continuous Numerical Values}

\author[1,$\ast$]{Brian P. Powell}
\author[1]{Jordan A Caraballo-Vega}
\author[1]{Mark L. Carroll}
\author[2,1]{Thomas Maxwell}
\author[1]{Andrew Ptak}
\author[3,1]{Greg Olmschenk}
\author[4,1]{Jorge Martinez-Palomera}
\affil[1]{NASA Goddard Space Flight Center, Greenbelt, MD 20771, USA}
\affil[2]{ARSC Federal Holding Company, Reston, VA 20190, USA}
\affil[3]{University of Maryland, College Park, MD 20742, USA}
\affil[4]{University of Maryland Baltimore County, Baltimore, MD 21250, USA}

\makeatletter
\g@addto@macro\@thanks{\footnotetext[1]{Corresponding author: \texttt{brian.p.powell@nasa.gov}}}
\makeatother

\date{}

\begin{document}

\maketitle

\begin{abstract}
We introduce Normalized Base-2 Encoding (NB2E) as a method for encoding continuous numerical values and demonstrate that, using this input encoding, vanilla multi-layer perceptrons (MLP) successfully extrapolate four periodic waveforms without prior knowledge of their functional form.  We contrast this finding with fixed Fourier encoding, which extrapolates well when the number of signal cycles over the normalized domain is an even integer and loses coherence at most other values, whereas NB2E degrades gradually away from those values and, except near odd cycle counts, remains substantially more accurate.  These findings are limited to a single MLP architecture trained on one synthetic, exactly periodic, one-dimensional signal at a time, with extrapolation extending only to the end of the normalized domain.  We do not propose a mechanism for this phenomenon. Rather, we present empirical observations which merit further study.
\end{abstract}

\section{Introduction}
\label{sec::intro}

Although machine learning (ML) methods are increasingly powerful, a model capable of true extrapolation has remained elusive. Neural networks are interpolators, not extrapolators. Indeed, while the Universal Approximation Theorem \citep[UAT;][]{cybenko1989approximation,hornik1989multilayer} suggests that a neural network can approximate any function in a bounded domain, it guarantees nothing outside of those bounds. In a broader sense, a model can only generalize to samples that are similar to the training data.
 
Without solving the extrapolation problem itself, various methods have been developed to effectively circumvent it. Among these is symbolic regression, which seeks to learn an equation representing the data \citep{martius2016extrapolation} or even the underlying physical laws \citep{schmidt2009distilling,udrescu2020ai}. While promising, these methods are typically applied to physical problems and are computationally expensive. Extrapolation is also not guaranteed, as the resulting symbolic expression is not necessarily correct beyond the training domain. Neural Ordinary Differential Equations (ODEs) \citep{chen2018neural} can extrapolate by modeling the dynamics of the data rather than the data itself. As such, they are readily applied to many types of data. However, extrapolation has a tendency to fail as errors accumulate in the dynamics \citep{rubanova2019latent,finlay2020how,ghosh2020steer}.
 
The most commonly used methods for prediction are sequence prediction models such as Recurrent Neural Networks (RNN), Gated Recurrent Units (GRU) \citep{cho2014learning}, Long Short-Term Memory (LSTM) \citep{hochreiter1997long,gers2000learning}, Temporal Convolutional Networks (TCN) \citep{oord2016wavenet,bai2018empirical}, Convolutional LSTM (ConvLSTM) \citep{shi2015convolutional} or other 3D CNN variants, Seq2Seq \citep{sutskever2014sequence}, and transformers \citep{vaswani2017attention}, among others. These methods all learn to predict an output sequence or value based on an input sequence. The sequence type can vary, with text data or time-series data being the most common. These types of models are ubiquitous. However, these models do not extrapolate beyond the training domain in the sense used here. They are better framed as autoregressive pattern matching, in which ``forecasting'' is prediction from previously seen sequences of learned representations \citep{bandara2020forecasting,lim2021timeseries}.

Conspicuously absent in a discussion of neural network extrapolation are Multi-Layer Perceptrons (MLPs), or basic feedforward neural networks.  This type of neural network was never meant to extrapolate a function.  A vanilla MLP is the most basic neural mechanism for approximating the value of a function within the bounds of its training data and, as expected, offers no predictive capability beyond those bounds.  Indeed, \citet{xu2021how} show that a Rectified Linear Unit (ReLU) MLP extrapolates linearly away from its training data regardless of the target.  In this paper, we investigate the ability of an input encoding to induce extrapolative capability in MLPs, specifically for periodic functions.  Since this type of network has no intrinsic extrapolative qualities, we focus on the input representation and use basic periodic signals as our test cases.

A stationary periodic signal can be extended by classical means that fit a sinusoidal model to samples and extend it indefinitely. We do not claim that an MLP can or needs to compete with these methods. The question here is narrower: what changing the input representation does to the behavior of a generic MLP beyond its training data. The comparisons in this paper are therefore between encodings of the same network, not between the network and classical estimators.

The choice of input representation fundamentally shapes what neural networks are capable of learning through inductive bias. For coordinate-based learning, various encoding methods have been developed to enable networks to capture high-frequency features.  Random Fourier features were proposed by \citet{rahimi2007} as a shift-invariant kernel method and adapted by \citet{tancik2020fourier} as a means of encoding for neural networks, whereby
\begin{equation}
\gamma(x) = [\cos(2\pi\omega_1 x), \sin(2\pi\omega_1 x), \cos(2\pi\omega_2 x), \sin(2\pi\omega_2 x), \ldots],
\label{eqn::fourier}
\end{equation}
where $\boldsymbol{\omega}_i \sim \mathcal{N}(\mathbf{0}, \sigma^2 \mathbf{I}), \quad i = 1, 2, \ldots, m$.

The random Fourier features of Equation \ref{eqn::fourier} were further developed for use in neural radiance fields \citep[NeRF,][]{mildenhall2020nerf} using deterministic features.  That is, for frequencies $\omega_i = 2^{i-1}$ where $i = 1, 2, \ldots, N$, fixed Fourier features are given by
\begin{equation}
\gamma(x) = [\sin(2\pi x), \cos(2\pi x), \sin(4\pi x), \cos(4\pi x), \sin(8\pi x), \cos(8\pi x), \ldots].
\label{eqn::ffe}
\end{equation}
Hereafter, to compare to our method and in recognition of its functional implementation as encoding coordinates, we will refer to this method as Fixed Fourier Encoding (FFE).

In the same paper that \citet{rahimi2007} introduced Fourier features, the authors also introduced random binning as a kernel method, with further analysis provided by \citet{rahimi2008weighted}. In random binning, the spatial domain is randomly partitioned (as opposed to the frequency domain with Fourier features), where a partition containing a given value is represented by unity and zero otherwise.  In this manner, it is representing a continuous value in a discrete space.  As FFE can be considered a deterministic form of random Fourier features, our concept can be considered a deterministic form of random spatial binning.  That is, the near-exact representation of a continuous value with a binary vector.  

This paper is structured as follows.  In Section \ref{sec::nb2e} we will introduce Normalized Base-2 Encoding (NB2E), with the limits thereof in Section \ref{sec::limit}.  Our results are shown in Section \ref{sec::results}, with subsections addressing (i) the comparison of NB2E to FFE and continuous input across encoding lengths and cycle counts, and (ii) sensitivity to the training fraction, sample count, and noise.  We provide a discussion in Section \ref{sec::discussion} and our conclusions in Section \ref{sec::conclusions}.

\section{Normalized Base-2 Encoding}
\label{sec::nb2e}

Binary encoding provides a fundamentally different representation than FFE. Both can be viewed as multi-resolution decompositions in that FFE provides continuous sinusoidal basis functions and binary encoding provides discrete step functions at hierarchical frequency scales. The discrete nature of binary representation may induce different learning dynamics in neural networks. We explore this through NB2E, which encodes continuous values in $[0,1)$ using their binary representation as a vector in the form
\begin{equation}
\gamma(x) = [B_1, B_2, \ldots, B_N], \quad \text{where } x = \sum_{i=1}^{N} B_i \times 2^{-i}, \; B_i \in \{0,1\},
\label{eqn::nb2e}
\end{equation}
where the components of the representation vector, viewed as functions of $x$, are the Rademacher functions \citep{rademacher1922einige} in $\{0,1\}$ form. 

\begin{figure}[h]
\centering
\includegraphics[width=0.4\textwidth]{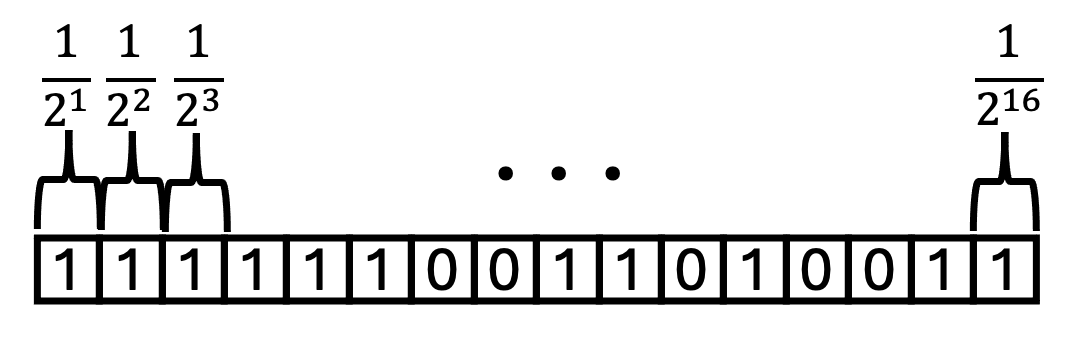}
\caption{16-element NB2E of the number 0.9876.  The given value is encoded into a vector of ones and zeros in the manner described by Equation \ref{eqn::nb2e} to be used as input to a neural network.}
\label{fig::encoding}
\end{figure}

NB2E implementation first requires the normalization of the input data to a $[0,1)$ scale, followed by an elementwise encoding of the base-2 representation. While normalization is not a strict requirement for binary encoding in general, we note that consistency of the meaning of each element of the binary vector is critical and normalization ensures that each continuous numerical value is encoded within the same context.  We also found normalization to be more amenable to managing the power of two boundaries of the encoding (discussed in Section \ref{sec::limit}).  In Figure \ref{fig::encoding}, we show an example NB2E representation of the number 0.9876.  

For all NB2E, the encoded vector of length $N$ has a representation given by Equation \ref{eqn::nb2e}. Clearly, the minimum NB2E representable value is 0, while the maximum representable value is $\sum_{i=1}^{N} 2^{-i}$, approaching a limit of unity with increasing N.  Represented values can naturally be more precise as $N$ increases.  Selection of $N$ should be customized to the needed precision of the application.\footnote{Bits finer than the spacing of the training samples carry no information about the target, and for FFE the highest octaves at $N=64$ exceed double precision. This is consistent with the small rise in error from $N=16$ to $N=64$ in Tables \ref{tab:results_N16_val} to \ref{tab:results_N64_val}.}  In Section \ref{sec::results}, we will detail the results of simulations with $N$ values of 16, 32, and 64. Any computational expense from varying $N$ is negligible, as the size of our neural network (described in Section \ref{sec::results}) remains the same, and the only modification to the number of parameters caused by changes in $N$ is in the weight and bias arrays between the input and the first layer.

\begin{figure}[h]
\centering
\includegraphics[width=1.0\textwidth]{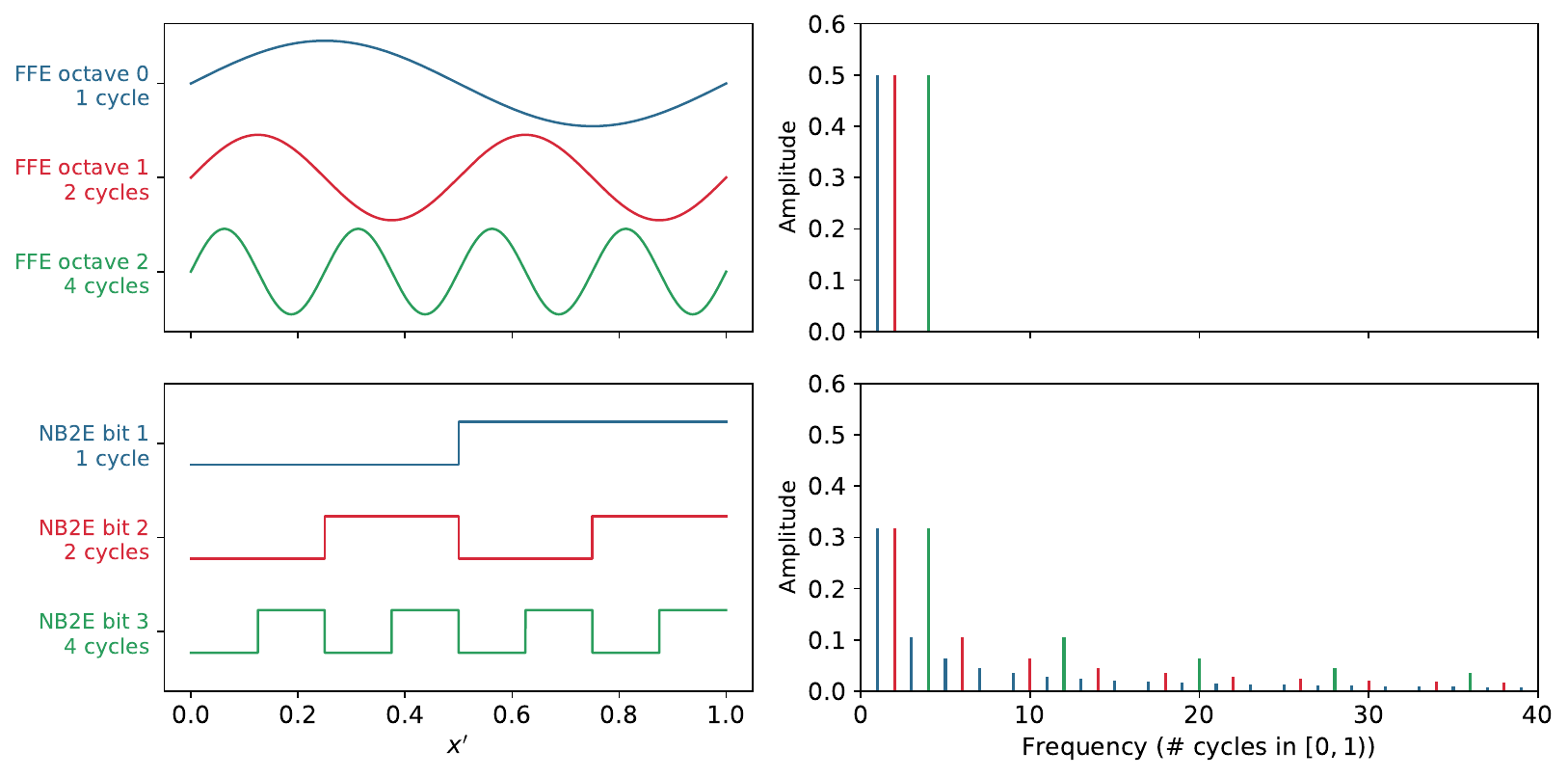}
\caption{For FFE (top panels) and NB2E (bottom panels) waveforms of the first three bits or octaves in the domain [0,1) are shown on the left and the frequency content of each waveform is shown on the right in corresponding colors.  Note that the sine and cosine components of FFE have the same frequency.}
\label{fig::freq}
\end{figure}

The NB2E representation consists of square waves at the frequency $f_i = 2^{i-1}$ of the bit index $i$.  The corresponding FFE consists of two values for each frequency, $\text{sin}(2\pi f_{i}x)$ and $\text{cos}(2\pi f_{i}x)$, with examples shown in Table \ref{tbl::encoding}.  We are careful to note, however, that the corresponding indices of the encoding do not contain identical frequency content.  Performing a Discrete Fourier Transform (DFT) on the waveform of each bit shows that the FFE contains only the given frequency, by identity, while the square wave contains additional frequency content.  We show this in Figure \ref{fig::freq}, where the left panels show the waveforms of the first three indices of the FFE and NB2E and the right panels show their equivalent frequency content obtained via DFT in corresponding colors.

\begin{table}[h]
\centering
\begin{tabular}{cccc}
\toprule
\textbf{Bit} & \textbf{Frequency} & \textbf{NB2E} & \textbf{FFE} \\
\midrule
$1$ & $2^{0} = 1$ & $2^{-1}$ & $\sin(2\pi x), \cos(2\pi x)$ \\
$2$ & $2^{1} = 2$ & $2^{-2}$ & $\sin(4\pi x), \cos(4\pi x)$ \\
$3$ & $2^{2} = 4$ & $2^{-3}$ & $\sin(8\pi x), \cos(8\pi x)$ \\
$4$ & $2^{3} = 8$ & $2^{-4}$ & $\sin(16\pi x), \cos(16\pi x)$ \\
$\vdots$ & $\vdots$ & $\vdots$ & $\vdots$ \\
$i$ & $2^{i-1}$ & $2^{-i}$ & $\sin(2^i\pi x), \cos(2^i\pi x)$ \\
\bottomrule
\end{tabular}
\caption{NB2E bits and the equivalent FFE. Each NB2E bit $i$ with positional value $2^{-i}$ corresponds to FFE at frequency $2^{i-1}$. The two encodings share a fundamental frequency at every index but differ in waveform as the square wave carries additional odd harmonics (see Figure \ref{fig::freq}).}
\label{tbl::encoding}
\end{table}

\section{NB2E Limits}
\label{sec::limit}

In order for a neural network to properly learn the meaning of the NB2E, each element of the encoded vector must have contained values of both zero and unity during training.  The first element of the NB2E vector (see Figure \ref{fig::encoding}) is $\frac{1}{2^1}=0.5$, so will not be unity for any number less than 0.5.  As such, the training data needs to extend beyond 0.5 in order to ensure the training of the first element of the encoding. That is, for input dataset $X$, normalized to $X'$ with subsets $X'_{\text{train}}$ and $X'_{\text{extrap}}$,
\begin{multline}
0 \leq \min(X'_{\text{train}}) < 0.5 < \max(X'_{\text{train}}) < \min(X'_{\text{extrap}}) < \max(X'_{\text{extrap}}) < 1.
\label{eqn::nb2e_ineq}
\end{multline}
We emphasize that $X'$ is normalized and therefore the size of the extrapolation domain scales with the domain of the non-normalized dataset.  

In order to expand the upper limit, effectively increasing the extrapolation horizon, normalization in NB2E need not be to the maximum value of the dataset.  That is, for dataset $X$ and normalizing value $z$, the normalized dataset $X'$ is obtained by the elementwise division

\begin{equation}
X' = \frac{X}{z}, \quad \text{where } z < 2\max(X),
\label{eqn::norm2}
\end{equation} 
such that $\max(X')>0.5$, abiding by Equation \ref{eqn::nb2e_ineq}.  This is an upper limit for $z$, however, and reasonable values should be much lower to ensure sufficient training.  For example, if a dataset is a time series for a duration of ten years and the desired extrapolation is two years into the future, simply divide the dataset by twelve years, i.e. $z=1.2\text{max}(X)$.  Train using $X'\leq\frac{5}{6}$, then extrapolate to any value $x'$ within $\frac{5}{6} < x' < 1$.  The extrapolation is then valid as a prediction for two years into the future, where $x=x'z$.  The outcome also depends on where the resulting cycle count falls, as Figure \ref{fig::sweep} shows. 

The normalizing value, $z$, should be carefully chosen and informed by the domain of the data and desired prediction domain, with consideration that Equations \ref{eqn::nb2e_ineq} and \ref{eqn::norm2} are limits rather than ideal values.  We will examine the sensitivity of the model predictions to the maximum value of the training domain in Section \ref{sec::sensitivity}, but will mention here that we suggest $\max(X'_{\text{train}})\geq0.7$ as a guiding principle for practical application.

\section{Results}
\label{sec::results}

\begin{figure}[h]
\centering
\includegraphics[width=0.5\textwidth]{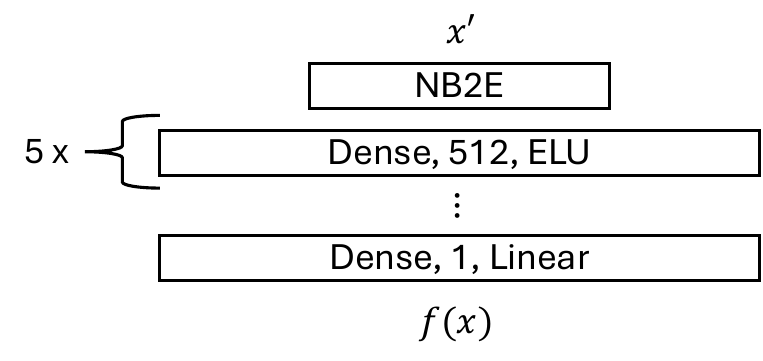}
\caption{MLP structure.  The NB2E is not part of the neural network, just noted as the encoding of the input.  Replace NB2E with FFE for the appropriate examples, whereas for continuous numerical input there is no encoding. There are five dense layers of 512 neurons each with ELU activation and $L_2$ regularization with a regularization factor of $10^{-4}$, followed by a dense layer with a single neuron and linear activation.}
\label{fig::nn}
\end{figure}

We tested the extrapolative ability of NB2E against FFE and continuous numerical input.  Our MLP (shown in Figure \ref{fig::nn}), built using \texttt{PyTorch} \citep{paszke2019pytorch}, consists of five dense layers containing 512 neurons each and using exponential linear unit (ELU) activation \citep{clevert2016fast}, with linear activation of a single neuron on the final layer.  $L_2$ regularization with a regularization factor of $10^{-4}$ is applied to each hidden layer to prevent overfitting, in addition to the decoupled weight decay of the optimizer described in Section \ref{sec::examples}.  Note that we do not use periodic activation functions \citep{parascandolo2016taming,sitzmann2020implicit} which, while they have not been claimed to allow for extrapolation, do provide internal representation of complex signals implicitly, whereas we wanted to test the qualities of the encoding directly.  

For each case in $X'_{\text{train}}$, the NB2E of $x'$ is given as input to the neural network, and $f(x)$ is given as a training target.  We emphasize the previous sentence for clarity, noting that $f(x)$ is the value of the function at the non-normalized $x$.  There are no dependencies trained as in sequence models, only the mapping from a single input to a single output per training sample.  

Training batches of length $B$ have inputs of shape $(B,N)$ for NB2E and $(B,2N)$ for FFE (which requires a sine and cosine term for each frequency, hence double the size of the NB2E, see Table \ref{tbl::encoding}), and $(B,1)$ for continuous numerical input.  Outputs, in all cases, are of shape $(B,1)$. We found batches of size $B=1000$ to train quickly and effectively.

\subsection{Examples}
\label{sec::examples}

\begin{figure}[h]
\centering
\includegraphics[width=1.0\textwidth]{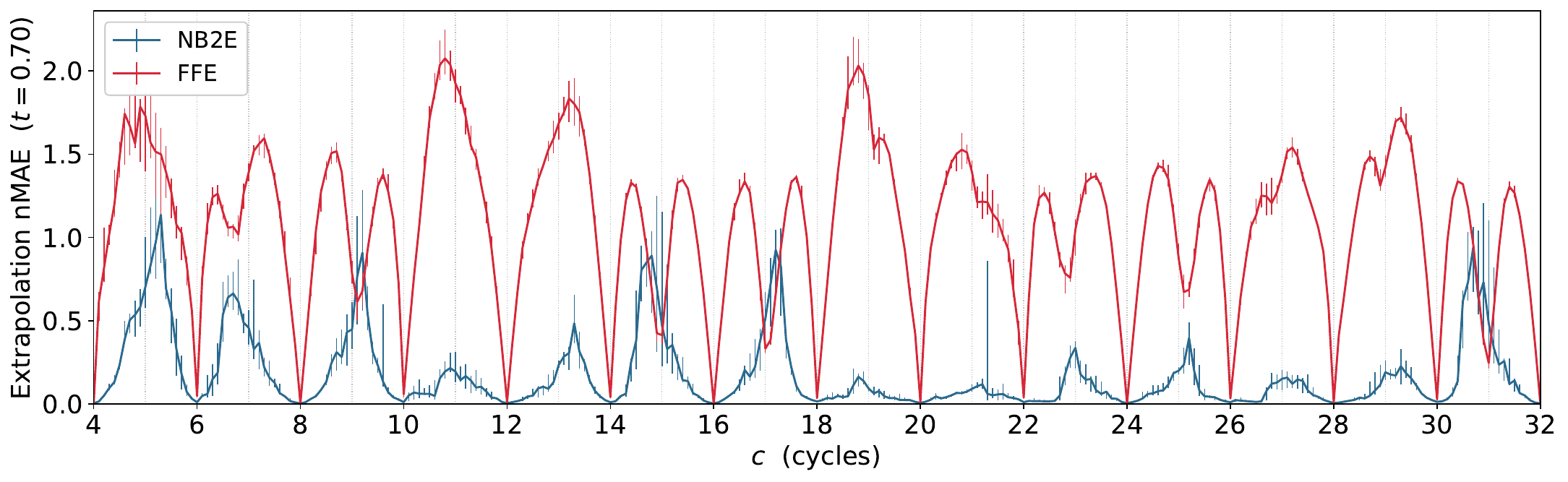}
\caption{Extrapolation nMAE for FFE (red) and NB2E (blue) from $c=4$ to $c=32$ in increments of 0.1. Error bars show the interquartile range over 15 runs (5 seeds $\times$ 3 phase offsets).}
\label{fig::sweep}
\end{figure}

We tested the neural network from Figure \ref{fig::nn} with NB2E, FFE, and continuous input for four different periodic functions, namely a sine, a triangle, a sawtooth, and a square wave. Each periodic function was tested at three cycle counts. The first is $c=24$, an even integer. The second is $c=2^{4.5}\approx22.63$, a non integer maximally distant from a power of two, which represents the general case and at which the NB2E error in Figure \ref{fig::sweep} is typical of non integer cycle counts in that range. The third is $c=17$, an odd integer adjacent to a power of two, near which NB2E has one of its largest errors in Figure \ref{fig::sweep} and at which FFE partially recovers.  For each of our tests, we used five different starting seeds and three different encoding lengths, $N$, of 16, 32, and 64. In all cases, we drew $\{x'_i\}_{i=1}^{10000} \overset{\text{i.i.d.}}{\sim} \mathcal{U}(0, 1)$ and used the function values at those samples below the training boundary, so that each training set contains approximately 7{,}000 points. We imposed randomness on $x'$ in order to avoid any suspicion that the MLP could learn patterns from regular intervals.

\begin{figure}[h]
\centering
\includegraphics[width=1.0\textwidth]{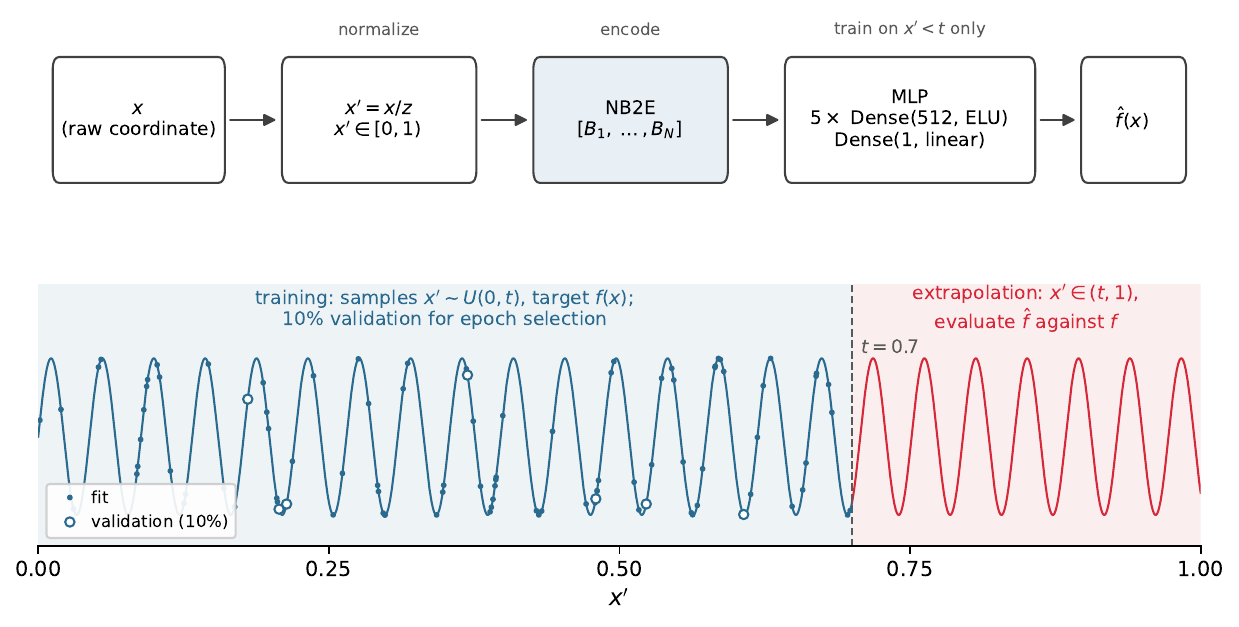}
\caption{A schematic of our process.  Raw coordinates, $x$, are normalized to $x'$ by the chosen normalization value, $z$.  $x'$ is then encoded into NB2E (top blue square). The MLP is then trained with the NB2E of $x'$ as the input and $f(x)$ as the output.  With $t$ as the boundary between the training region and extrapolation region, we train the MLP on all $x'<t$, with 10\% of the training data reserved as validation data for selection of the best training epoch.  The best weights are then used to predict the value of the function in the extrapolation region, $(t,1)$.}
\label{fig::schema}
\end{figure}

The normalized samples with $x'_i < t$ form the training set $X'_{\text{train}}$, where hereafter $t$ denotes this boundary, and the extrapolation region is $X'_{\text{extrap}} = (t, 1)$. We fix the boundary at $t = 0.7$ rather than the sample count so that every configuration is assessed on the same domain. Errors in both regions are evaluated on a dense grid that is independent of the samples. The grid has 4{,}000 evenly spaced points per region with additional points placed within $10^{-4}$ of every bit-transition boundary to depth 7. We provide a schematic of our process in Figure \ref{fig::schema}.

We used the AdamW optimizer \citep{kingma2015adam,loshchilov2019decoupled} with decoupled weight decay of $4\times10^{-3}$ and a cosine annealing learning rate scheduler with warm restarts \citep{loshchilov2017sgdr}\footnote{Warm restarts reset the learning rate to a higher value at a fixed schedule during training, which causes reduced performance temporarily while the model seeks a new minimum.  While not optimal for all applications, this method can improve outcomes in the presence of false local minima.}. The learning rate begins at $10^{-3}$ and decays to $10^{-5}$ over each cycle, where the first cycle is 100 optimizer steps and each subsequent cycle is twice as long as the one before it. The Mean Absolute Error (MAE) was our objective function in all cases, and every model was trained for 1{,}500 epochs. A random 10\% of the training samples is set aside as a validation subset and is not used for fitting. The weights retained are those of the epoch with the lowest MAE on the validation subset, evaluated at every epoch. The extrapolation region is never used for selection. Five random seeds are trained per configuration, each drawing its own samples, validation subset, and initialization. On a single NVIDIA Tesla V100 GPU, training completes in 1 to 3 minutes.

We report the normalized mean absolute error,
\begin{equation}
  \mathrm{nMAE} \;=\; \frac{\overline{\left|\hat f - f\right|}}{\overline{\left|f - \bar f\right|}},
  \label{eqn::nmae}
\end{equation}
where $\hat f$ is the prediction, $f$ the clean target, and $\bar f$ its mean. The overline denotes the mean over evaluation points, taken over the region being reported in the numerator and over the full domain $[0,1)$ in the denominator, so that the denominator is the mean absolute deviation of the target function and is the same constant for every region and encoding. An nMAE of 1 corresponds to predicting the mean.

\begin{figure}[h]
\centering
\includegraphics[width=1.0\textwidth]{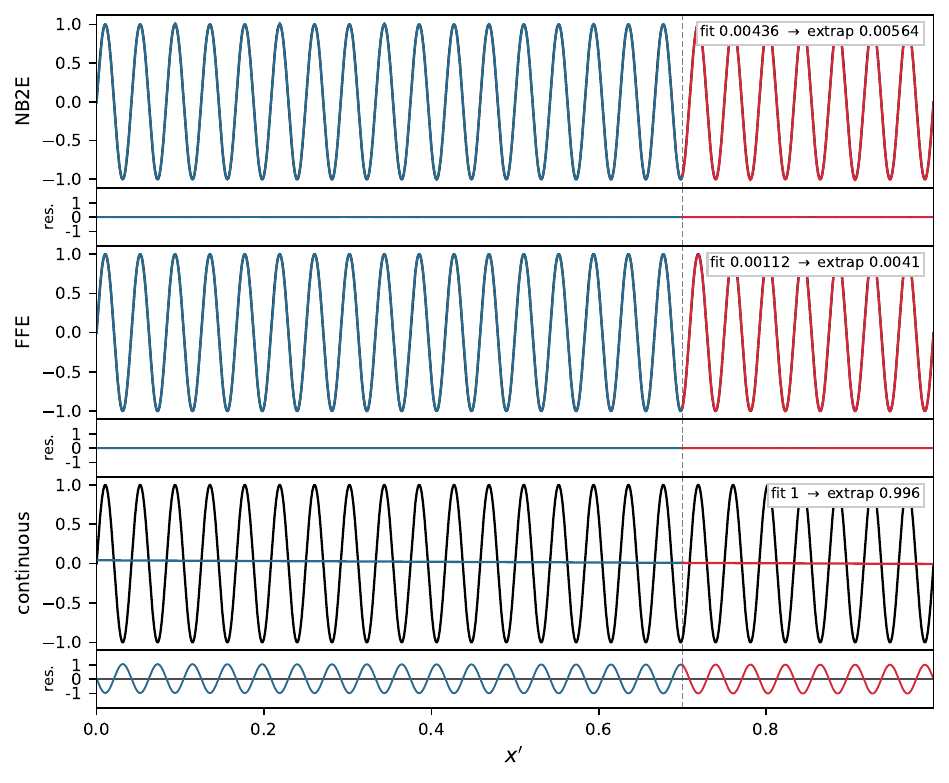}
\caption{Results of training the MLP from Figure \ref{fig::nn} with encoded input data for $f(x') = \sin(2\pi \cdot 24 x')$.  Predictions for the training data (blue) and the extrapolation region (red) for ({\em top}) NB2E, ({\em middle}) FFE, and ({\em bottom}) continuous, with residuals plotted on the same scale for visual comparison.  The true function is in black.  Fit and extrapolation nMAE for each case are given in the text boxes in each panel.}
\label{fig::sine24}
\end{figure}

We will first examine the results of the test of the sine function in Figure \ref{fig::sine24} in order to clearly demonstrate our process.  The results shown in the figure are specifically those of the median error seed at $N=16$ and $c=24$.  The predictions of the training data are in blue and the extrapolation region in red.  The top panel shows NB2E, the middle panel shows FFE, and the bottom panel shows continuous input for comparison. Residuals are under their respective panels.  The nMAE for both the fit and extrapolation are shown in a text box in each panel.  We emphasize that during training, the MLP never saw the encoding of any input value greater than $t=0.7$. It can be seen in this figure that both the NB2E and FFE fit the sine function very nicely, with the FFE showing the best nMAE results for both the fit and the extrapolation.  Continuous input could not even fit the function in training, consistent with the spectral bias of scalar input networks \citep{rahaman2019spectral}, which demonstrates the induced learning from both FFE and NB2E.  In all cases tested in this work, we emphasize that continuous input failed to fit the periodic function in the training region, and the extrapolation performance of continuous input should therefore be considered as a comparison against a naive extrapolator rather than a competitive baseline.

\begin{figure}[h]
\centering
\includegraphics[width=1.0\textwidth]{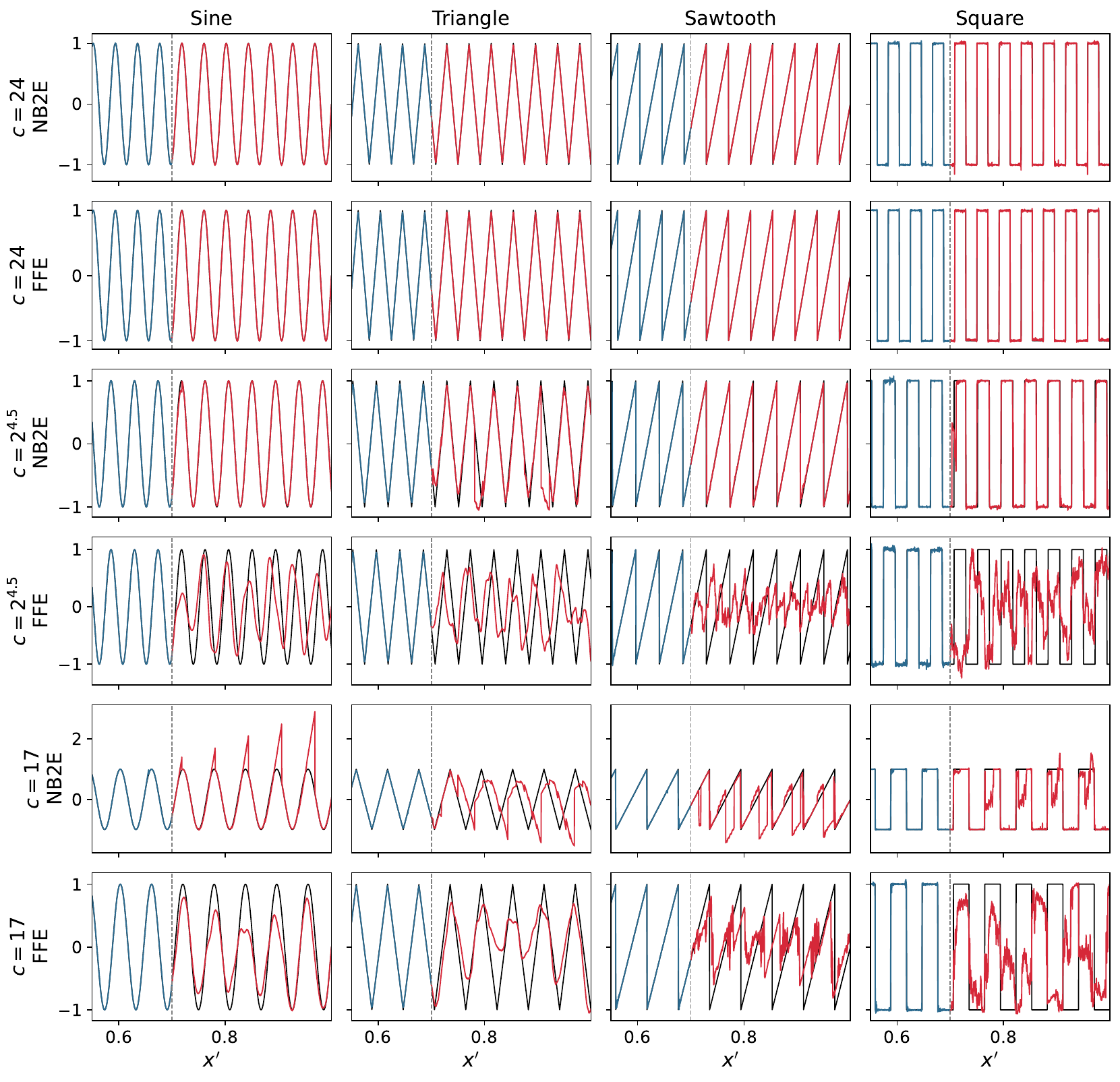}
\caption{Extrapolation results for the sine, triangle, sawtooth, and square functions at $c=24$ (top two rows), $c=2^{4.5}$ (middle two rows), and $c=17$ (bottom two rows) at $N=16$, for NB2E and FFE. Each panel shows the seed with the median extrapolation error. All $c=24$ examples extrapolate well for both FFE and NB2E, only NB2E extrapolation remains coherent when $c=2^{4.5}$, and both encodings are degraded at $c=17$. The NB2E triangle at $c=2^{4.5}$ varies substantially across seeds (see Table \ref{tab:results_N16_val}).}
\label{fig::ex}
\end{figure}

The results at $c=2^{4.5}$ with all other parameters identical to the previous test, shown in the middle rows of Figure \ref{fig::ex}, are instructive.  Here the FFE is unable to extrapolate the signal coherently, whereas NB2E extrapolation remains quite good.  Figure \ref{fig::ex} shows the same comparison for all four periodic functions at all three cycle counts.  We provide detailed results in Table \ref{tab:results_N16_val}, Table \ref{tab:results_N32_val}, and Table \ref{tab:results_N64_val} for $N=16$, $N=32$, and $N=64$, respectively.  The training nMAE is provided in the upper section of each of the tables, with the extrapolation nMAE in the lower section.  Each reported value is the median of five starting seeds, with the range provided in parentheses. The lowest nMAE for each function/$c$ combination is given in bold in each row.  The pattern is clear: (i) Both FFE and NB2E fit the training data nicely, with FFE slightly better in most cases.  (ii)  Both FFE and NB2E successfully extrapolate all $c=24$ cases with FFE again performing slightly better.  (iii) At $c=2^{4.5}$, FFE fails to extrapolate on every waveform while NB2E extrapolates all four. The triangle function is the weakest and most seed-dependent NB2E case, with a range of 0.03 to 0.50 at $N=16$ as shown in Table \ref{tab:results_N16_val}. (iv) At $c=17$, both encodings are degraded on every waveform and vary widely across seeds, as Figure \ref{fig::ex} and Tables \ref{tab:results_N16_val} to \ref{tab:results_N64_val} show. NB2E is far better than FFE on the square, FFE is better on the triangle, where NB2E fails outright at $N=16$, the two are comparable on the sine at $N=16$ and $N=32$ with NB2E better at $N=64$, and they are comparable on the sawtooth at $N=16$ with NB2E better at the larger lengths.

\begin{figure}[h]
\centering
\includegraphics[width=1.0\textwidth]{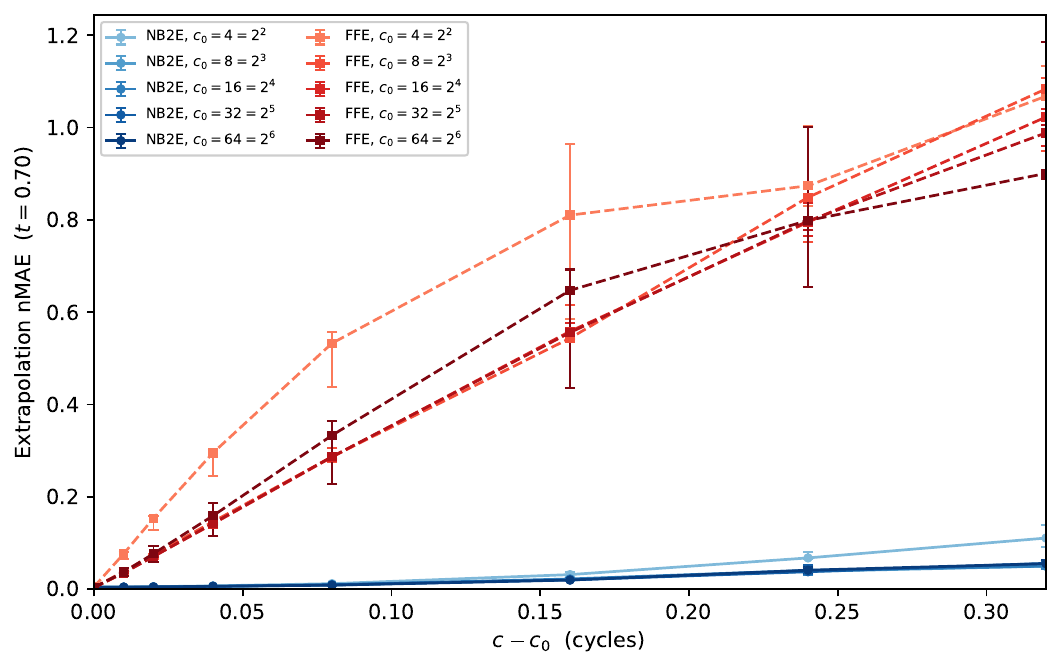}
\caption{Extrapolation nMAE as a function of distance in $c$ from a power of two, for $c_0=\{4,8,16,32,64\}$.  NB2E is shown in shades of blue, while FFE is shown in shades of red.  Points are the median nMAE and error bars show the interquartile range over 15 runs (5 seeds $\times$ 3 phase offsets).}
\label{fig::pow}
\end{figure}

It is clear that the dominant feature driving performance is the number of cycles, $c$, over the normalized domain, and three cycle counts are not enough to characterize it. We therefore ran two further experiments with the sine only, at $N=16$, for five seeds and three phase shifts $\phi \in \{0, 2.33, 4.19\}$ rad on $f(x')=\sin(2\pi cx'+\phi)$, the phase shifts being a hedge against a zero phase target being accidentally beneficial to either encoding. First, at $c_0 \in \{4, 8, 16, 32, 64\}$ we set $c = c_0 + \varepsilon$ for $\varepsilon \in \{0, 0.01, 0.02, 0.04, 0.08, 0.16, 0.24, 0.32\}$ (Figure \ref{fig::pow}). Second, we swept $c$ from 4 to 32 in steps of 0.1 (Figure \ref{fig::sweep}). In both figures each point is the median of the 15 runs and the error bars are the interquartile range. We used the sine only because Figure \ref{fig::ex} and Tables \ref{tab:results_N16_val} to \ref{tab:results_N64_val} show the four waveforms and three encoding lengths behaving alike at $c=24$ and $c=2^{4.5}$, which is the behavior the sweep is meant to generalize. At $c=17$ the waveforms diverge, so the sweep characterizes the sine.

In Figure \ref{fig::pow}, we show the nMAE of the FFE and NB2E as a function of distance of $c$ from a power of two.  Both encodings are dyadic in nature, which is why we anchored this experiment at powers of two, though Figure \ref{fig::sweep} shows that every even integer behaves the same way and the predictor introduced below makes the same prediction at any even anchor.  Figure \ref{fig::pow} should therefore be read as the local behavior near five even anchors. It is particularly compelling that the nMAE of the FFE case increases sharply as $c$ drifts from the anchor, whereas the nMAE of the NB2E case rises gradually and to a much lower level.

In Figure \ref{fig::sweep}, we show the extrapolation nMAE of FFE and NB2E from $c=4$ to $c=32$ in increments of 0.1, with FFE shown in red and NB2E shown in blue.  It clearly shows that, rather than only at powers of two, $c$ at all even integers allows for excellent extrapolation results with both FFE and NB2E.  The nMAE for both FFE and NB2E increases between even integers, with the FFE nMAE tending to be substantially larger than NB2E, although that is not universally true. The largest NB2E errors lie in narrow bands near, rather than at, the odd integers adjacent to powers of two, reaching 0.9 to 1.1 near $c = 5.3$, 9.2, 14.8, 17.2, and 30.7, and FFE partially recovers at several odd integers, to about 0.25 to 0.4 at $c = 15$, 17, and 31. At $c = 15$ and 17 the two encodings are comparable, at $c = 31$ FFE is better, and in the bands just above 9, 15, and 17, where NB2E reaches about 0.9 and FFE about 0.7, FFE is the better of the two.\footnote{The isolated NB2E spike near $c = 21.3$ is four of five seeds at one phase offset, with the other two offsets and the neighboring cycle counts unaffected. It is an instance of the phase sensitivity the offsets were included to expose.}  

A simple predictor makes the even integer result and the training fraction limit precise, and also shows what the NB2E result is not. Every FFE feature above octave 0 and every NB2E bit from bit 2 onward is unchanged by a shift of $x'$ by one half, so a network that disregards the coarsest feature outputs $f(x' - \tfrac{1}{2})$ throughout the extrapolation region, a shifted copy of its own fit to the training region. The copy is exact when $c$ is an even integer and otherwise carries a phase error of $c/2$ cycles, whose fractional part is $\varepsilon/2$ for $c = c_0 + \varepsilon$ with $c_0$ even, giving nMAE $2\sin(\pi\varepsilon/2)$. The FFE error in Figure \ref{fig::pow} tracks this prediction within about 20 percent at every offset for $c_0 = 8$, 16, and 32, exceeds it by up to a third at $c_0 = 64$, and is about twice the prediction at small offsets for $c_0 = 4$. Every FFE point lies above the prediction, so FFE behaves as such a copy and a little worse. At $t=0.5$ in Figure \ref{fig::vsplit} the NB2E output is such a copy, with nMAE 1.65 against the predicted 1.66. At an odd integer the copy would carry a half cycle error and score 2, and both encodings do far better than that at $c = 17$, so the odd integer degradation is not the copy either. Away from even integers, however, neither network is a shifted copy. Comparing each trained network's output on $(t,1)$ with its own output on $(t - \tfrac{1}{2}, \tfrac{1}{2})$ gives an nMAE of about 0.004 at $c=24$ for both encodings, where the two are indistinguishable from the true function, but 1.6 for both encodings at $c=2^{4.5}$. There the copy would give 1.66, FFE gives 0.97, and NB2E gives 0.018. The NB2E result away from even integers is therefore neither a shifted copy of the training data nor a behavior it shares with FFE, and we do not know what produces it. This comparison is the reason we regard that case as the result of the paper.

\begin{table}[p]
\centering
\footnotesize
\caption{Training and extrapolation error (nMAE) at encoding width $N=16$ (16 bits for NB2E, 16 octaves for FFE) for the three cycle counts of Section \ref{sec::examples}. The weights retained are those of the epoch with the lowest MAE on the validation subset. The median over five seeds is provided with the (min - max) range for each value. The best result for each row is in bold. All tests share the same MLP, optimizer, starting seeds, and epoch budget.}
\label{tab:results_N16_val}
\begin{tabular}{lrrr}
\toprule
 & \multicolumn{3}{c}{Training} \\
\cmidrule(lr){2-4}
Function & NB2E & FFE & Continuous \\
\midrule
sine, $c=24$ & 0.004 {\scriptsize(0.003--0.005)} & \textbf{0.001 {\scriptsize(0.001--0.001)}} & 1.003 {\scriptsize(1.003--1.004)} \\
triangle, $c=24$ & 0.006 {\scriptsize(0.005--0.006)} & \textbf{0.003 {\scriptsize(0.002--0.003)}} & 0.995 {\scriptsize(0.995--0.997)} \\
sawtooth, $c=24$ & 0.012 {\scriptsize(0.011--0.015)} & \textbf{0.007 {\scriptsize(0.007--0.008)}} & 1.007 {\scriptsize(1.006--1.008)} \\
square, $c=24$ & 0.007 {\scriptsize(0.006--0.008)} & \textbf{0.004 {\scriptsize(0.004--0.005)}} & 0.987 {\scriptsize(0.986--0.999)} \\
\addlinespace
sine, $c=2^{4.5}$ & 0.006 {\scriptsize(0.005--0.006)} & \textbf{0.002 {\scriptsize(0.002--0.002)}} & 1.005 {\scriptsize(0.998--1.007)} \\
triangle, $c=2^{4.5}$ & 0.014 {\scriptsize(0.013--0.016)} & \textbf{0.008 {\scriptsize(0.008--0.008)}} & 0.991 {\scriptsize(0.991--0.993)} \\
sawtooth, $c=2^{4.5}$ & 0.029 {\scriptsize(0.027--0.030)} & 0.029 {\scriptsize(0.027--0.030)} & 0.999 {\scriptsize(0.999--1.000)} \\
square, $c=2^{4.5}$ & \textbf{0.019 {\scriptsize(0.016--0.023)}} & 0.027 {\scriptsize(0.026--0.028)} & 0.981 {\scriptsize(0.950--0.993)} \\
\addlinespace
sine, $c=17$ & 0.006 {\scriptsize(0.005--0.006)} & \textbf{0.002 {\scriptsize(0.002--0.002)}} & 1.005 {\scriptsize(0.957--1.007)} \\
triangle, $c=17$ & 0.012 {\scriptsize(0.011--0.014)} & \textbf{0.006 {\scriptsize(0.005--0.006)}} & 0.992 {\scriptsize(0.992--0.997)} \\
sawtooth, $c=17$ & 0.021 {\scriptsize(0.018--0.022)} & \textbf{0.018 {\scriptsize(0.017--0.018)}} & 1.006 {\scriptsize(1.005--1.008)} \\
square, $c=17$ & \textbf{0.011 {\scriptsize(0.010--0.013)}} & 0.014 {\scriptsize(0.012--0.015)} & 0.936 {\scriptsize(0.788--0.995)} \\
\midrule
 & \multicolumn{3}{c}{Extrapolation} \\
\cmidrule(lr){2-4}
Function & NB2E & FFE & Continuous \\
\midrule
sine, $c=24$ & 0.006 {\scriptsize(0.004--0.006)} & \textbf{0.004 {\scriptsize(0.002--0.006)}} & 0.996 {\scriptsize(0.995--0.997)} \\
triangle, $c=24$ & 0.007 {\scriptsize(0.006--0.008)} & \textbf{0.005 {\scriptsize(0.004--0.006)}} & 1.013 {\scriptsize(1.006--1.037)} \\
sawtooth, $c=24$ & 0.015 {\scriptsize(0.015--0.025)} & \textbf{0.011 {\scriptsize(0.010--0.012)}} & 0.992 {\scriptsize(0.990--1.002)} \\
square, $c=24$ & 0.010 {\scriptsize(0.008--0.012)} & \textbf{0.009 {\scriptsize(0.006--0.010)}} & 1.092 {\scriptsize(0.962--1.415)} \\
\addlinespace
sine, $c=2^{4.5}$ & \textbf{0.018 {\scriptsize(0.015--0.045)}} & 0.972 {\scriptsize(0.867--1.025)} & 0.994 {\scriptsize(0.989--1.027)} \\
triangle, $c=2^{4.5}$ & \textbf{0.285 {\scriptsize(0.032--0.502)}} & 1.150 {\scriptsize(1.108--1.270)} & 1.011 {\scriptsize(1.011--1.016)} \\
sawtooth, $c=2^{4.5}$ & \textbf{0.058 {\scriptsize(0.044--0.069)}} & 0.980 {\scriptsize(0.878--1.080)} & 1.013 {\scriptsize(1.000--1.031)} \\
square, $c=2^{4.5}$ & \textbf{0.076 {\scriptsize(0.065--0.083)}} & 0.875 {\scriptsize(0.869--0.909)} & 1.279 {\scriptsize(0.972--3.043)} \\
\addlinespace
sine, $c=17$ & \textbf{0.293 {\scriptsize(0.256--0.311)}} & 0.353 {\scriptsize(0.328--0.507)} & 0.990 {\scriptsize(0.985--1.009)} \\
triangle, $c=17$ & 1.204 {\scriptsize(0.617--1.455)} & \textbf{0.599 {\scriptsize(0.495--0.643)}} & 1.013 {\scriptsize(1.011--1.024)} \\
sawtooth, $c=17$ & 0.897 {\scriptsize(0.813--1.219)} & \textbf{0.890 {\scriptsize(0.854--1.181)}} & 0.989 {\scriptsize(0.988--1.008)} \\
square, $c=17$ & \textbf{0.194 {\scriptsize(0.082--0.391)}} & 1.108 {\scriptsize(1.072--1.209)} & 0.951 {\scriptsize(0.948--1.869)} \\
\bottomrule
\end{tabular}
\end{table}

\begin{table}[p]
\centering
\footnotesize
\caption{Same as Table \ref{tab:results_N16_val} for $N=32$.}
\label{tab:results_N32_val}
\begin{tabular}{lrrr}
\toprule
 & \multicolumn{3}{c}{Training} \\
\cmidrule(lr){2-4}
Function & NB2E & FFE & Continuous \\
\midrule
sine, $c=24$ & 0.006 {\scriptsize(0.006--0.007)} & \textbf{0.002 {\scriptsize(0.002--0.002)}} & 1.003 {\scriptsize(1.003--1.004)} \\
triangle, $c=24$ & 0.009 {\scriptsize(0.008--0.009)} & \textbf{0.004 {\scriptsize(0.004--0.004)}} & 0.995 {\scriptsize(0.995--0.997)} \\
sawtooth, $c=24$ & 0.016 {\scriptsize(0.014--0.017)} & \textbf{0.010 {\scriptsize(0.009--0.011)}} & 1.007 {\scriptsize(1.006--1.008)} \\
square, $c=24$ & 0.008 {\scriptsize(0.007--0.011)} & \textbf{0.006 {\scriptsize(0.005--0.006)}} & 0.987 {\scriptsize(0.986--0.999)} \\
\addlinespace
sine, $c=2^{4.5}$ & 0.011 {\scriptsize(0.010--0.012)} & \textbf{0.004 {\scriptsize(0.003--0.005)}} & 1.005 {\scriptsize(0.998--1.007)} \\
triangle, $c=2^{4.5}$ & 0.019 {\scriptsize(0.018--0.021)} & \textbf{0.011 {\scriptsize(0.011--0.012)}} & 0.991 {\scriptsize(0.991--0.993)} \\
sawtooth, $c=2^{4.5}$ & \textbf{0.041 {\scriptsize(0.037--0.045)}} & 0.045 {\scriptsize(0.042--0.046)} & 0.999 {\scriptsize(0.999--1.000)} \\
square, $c=2^{4.5}$ & \textbf{0.032 {\scriptsize(0.031--0.035)}} & 0.043 {\scriptsize(0.040--0.044)} & 0.981 {\scriptsize(0.950--0.993)} \\
\addlinespace
sine, $c=17$ & 0.010 {\scriptsize(0.010--0.011)} & \textbf{0.004 {\scriptsize(0.004--0.004)}} & 1.005 {\scriptsize(0.957--1.007)} \\
triangle, $c=17$ & 0.015 {\scriptsize(0.014--0.016)} & \textbf{0.009 {\scriptsize(0.009--0.010)}} & 0.992 {\scriptsize(0.992--0.997)} \\
sawtooth, $c=17$ & 0.032 {\scriptsize(0.030--0.035)} & 0.032 {\scriptsize(0.029--0.034)} & 1.006 {\scriptsize(1.005--1.008)} \\
square, $c=17$ & 0.021 {\scriptsize(0.020--0.023)} & 0.021 {\scriptsize(0.019--0.023)} & 0.936 {\scriptsize(0.788--0.995)} \\
\midrule
 & \multicolumn{3}{c}{Extrapolation} \\
\cmidrule(lr){2-4}
Function & NB2E & FFE & Continuous \\
\midrule
sine, $c=24$ & 0.007 {\scriptsize(0.006--0.007)} & \textbf{0.004 {\scriptsize(0.003--0.006)}} & 0.996 {\scriptsize(0.995--0.997)} \\
triangle, $c=24$ & 0.010 {\scriptsize(0.009--0.010)} & \textbf{0.005 {\scriptsize(0.005--0.005)}} & 1.013 {\scriptsize(1.006--1.037)} \\
sawtooth, $c=24$ & 0.021 {\scriptsize(0.020--0.025)} & \textbf{0.013 {\scriptsize(0.011--0.015)}} & 0.992 {\scriptsize(0.990--1.002)} \\
square, $c=24$ & 0.011 {\scriptsize(0.009--0.014)} & \textbf{0.009 {\scriptsize(0.008--0.010)}} & 1.092 {\scriptsize(0.962--1.415)} \\
\addlinespace
sine, $c=2^{4.5}$ & \textbf{0.015 {\scriptsize(0.014--0.017)}} & 0.998 {\scriptsize(0.903--1.038)} & 0.994 {\scriptsize(0.989--1.027)} \\
triangle, $c=2^{4.5}$ & \textbf{0.170 {\scriptsize(0.024--0.394)}} & 1.150 {\scriptsize(1.021--1.208)} & 1.011 {\scriptsize(1.011--1.016)} \\
sawtooth, $c=2^{4.5}$ & \textbf{0.063 {\scriptsize(0.059--0.070)}} & 0.980 {\scriptsize(0.891--1.107)} & 1.013 {\scriptsize(1.000--1.031)} \\
square, $c=2^{4.5}$ & \textbf{0.083 {\scriptsize(0.075--0.088)}} & 1.029 {\scriptsize(0.878--1.057)} & 1.279 {\scriptsize(0.972--3.043)} \\
\addlinespace
sine, $c=17$ & \textbf{0.305 {\scriptsize(0.286--0.314)}} & 0.425 {\scriptsize(0.328--0.479)} & 0.990 {\scriptsize(0.985--1.009)} \\
triangle, $c=17$ & 0.796 {\scriptsize(0.550--0.900)} & \textbf{0.566 {\scriptsize(0.499--0.692)}} & 1.013 {\scriptsize(1.011--1.024)} \\
sawtooth, $c=17$ & \textbf{0.590 {\scriptsize(0.240--0.658)}} & 0.942 {\scriptsize(0.845--1.106)} & 0.989 {\scriptsize(0.988--1.008)} \\
square, $c=17$ & \textbf{0.107 {\scriptsize(0.088--0.580)}} & 1.090 {\scriptsize(1.064--1.212)} & 0.951 {\scriptsize(0.948--1.869)} \\
\bottomrule
\end{tabular}
\end{table}

\begin{table}[p]
\centering
\footnotesize
\caption{Same as Table \ref{tab:results_N16_val} for $N=64$.}
\label{tab:results_N64_val}
\begin{tabular}{lrrr}
\toprule
 & \multicolumn{3}{c}{Training} \\
\cmidrule(lr){2-4}
Function & NB2E & FFE & Continuous \\
\midrule
sine, $c=24$ & 0.008 {\scriptsize(0.007--0.009)} & \textbf{0.003 {\scriptsize(0.003--0.003)}} & 1.003 {\scriptsize(1.003--1.004)} \\
triangle, $c=24$ & 0.011 {\scriptsize(0.010--0.011)} & \textbf{0.006 {\scriptsize(0.006--0.007)}} & 0.995 {\scriptsize(0.995--0.997)} \\
sawtooth, $c=24$ & 0.023 {\scriptsize(0.022--0.026)} & \textbf{0.018 {\scriptsize(0.015--0.020)}} & 1.007 {\scriptsize(1.006--1.008)} \\
square, $c=24$ & 0.012 {\scriptsize(0.011--0.015)} & \textbf{0.008 {\scriptsize(0.007--0.008)}} & 0.987 {\scriptsize(0.986--0.999)} \\
\addlinespace
sine, $c=2^{4.5}$ & 0.016 {\scriptsize(0.015--0.018)} & \textbf{0.010 {\scriptsize(0.008--0.012)}} & 1.005 {\scriptsize(0.998--1.007)} \\
triangle, $c=2^{4.5}$ & 0.023 {\scriptsize(0.021--0.024)} & \textbf{0.017 {\scriptsize(0.017--0.018)}} & 0.991 {\scriptsize(0.991--0.993)} \\
sawtooth, $c=2^{4.5}$ & \textbf{0.064 {\scriptsize(0.063--0.072)}} & 0.074 {\scriptsize(0.071--0.077)} & 0.999 {\scriptsize(0.999--1.000)} \\
square, $c=2^{4.5}$ & \textbf{0.047 {\scriptsize(0.044--0.053)}} & 0.065 {\scriptsize(0.061--0.067)} & 0.981 {\scriptsize(0.950--0.993)} \\
\addlinespace
sine, $c=17$ & 0.015 {\scriptsize(0.014--0.016)} & \textbf{0.008 {\scriptsize(0.007--0.008)}} & 1.005 {\scriptsize(0.957--1.007)} \\
triangle, $c=17$ & 0.017 {\scriptsize(0.014--0.018)} & \textbf{0.016 {\scriptsize(0.016--0.018)}} & 0.992 {\scriptsize(0.992--0.997)} \\
sawtooth, $c=17$ & 0.066 {\scriptsize(0.048--0.070)} & \textbf{0.065 {\scriptsize(0.058--0.065)}} & 1.006 {\scriptsize(1.005--1.008)} \\
square, $c=17$ & 0.042 {\scriptsize(0.038--0.050)} & \textbf{0.038 {\scriptsize(0.036--0.049)}} & 0.936 {\scriptsize(0.788--0.995)} \\
\midrule
 & \multicolumn{3}{c}{Extrapolation} \\
\cmidrule(lr){2-4}
Function & NB2E & FFE & Continuous \\
\midrule
sine, $c=24$ & 0.008 {\scriptsize(0.008--0.009)} & \textbf{0.005 {\scriptsize(0.004--0.008)}} & 0.996 {\scriptsize(0.995--0.997)} \\
triangle, $c=24$ & 0.011 {\scriptsize(0.010--0.011)} & \textbf{0.008 {\scriptsize(0.007--0.009)}} & 1.013 {\scriptsize(1.006--1.037)} \\
sawtooth, $c=24$ & 0.034 {\scriptsize(0.030--0.037)} & \textbf{0.019 {\scriptsize(0.019--0.023)}} & 0.992 {\scriptsize(0.990--1.002)} \\
square, $c=24$ & 0.017 {\scriptsize(0.016--0.022)} & \textbf{0.011 {\scriptsize(0.010--0.011)}} & 1.092 {\scriptsize(0.962--1.415)} \\
\addlinespace
sine, $c=2^{4.5}$ & \textbf{0.017 {\scriptsize(0.015--0.019)}} & 0.947 {\scriptsize(0.877--0.998)} & 0.994 {\scriptsize(0.989--1.027)} \\
triangle, $c=2^{4.5}$ & \textbf{0.037 {\scriptsize(0.025--0.239)}} & 1.073 {\scriptsize(1.051--1.245)} & 1.011 {\scriptsize(1.011--1.016)} \\
sawtooth, $c=2^{4.5}$ & \textbf{0.079 {\scriptsize(0.073--0.101)}} & 1.113 {\scriptsize(0.976--1.143)} & 1.013 {\scriptsize(1.000--1.031)} \\
square, $c=2^{4.5}$ & \textbf{0.086 {\scriptsize(0.083--0.095)}} & 0.912 {\scriptsize(0.852--1.041)} & 1.279 {\scriptsize(0.972--3.043)} \\
\addlinespace
sine, $c=17$ & \textbf{0.269 {\scriptsize(0.213--0.316)}} & 0.683 {\scriptsize(0.533--0.840)} & 0.990 {\scriptsize(0.985--1.009)} \\
triangle, $c=17$ & 0.729 {\scriptsize(0.631--1.259)} & \textbf{0.627 {\scriptsize(0.552--0.720)}} & 1.013 {\scriptsize(1.011--1.024)} \\
sawtooth, $c=17$ & \textbf{0.229 {\scriptsize(0.173--0.643)}} & 0.875 {\scriptsize(0.833--0.965)} & 0.989 {\scriptsize(0.988--1.008)} \\
square, $c=17$ & \textbf{0.170 {\scriptsize(0.133--1.148)}} & 1.110 {\scriptsize(1.079--1.143)} & 0.951 {\scriptsize(0.948--1.869)} \\
\bottomrule
\end{tabular}
\end{table}

\subsection{Sensitivity Analysis}
\label{sec::sensitivity}

\begin{figure}[h]
\centering

\includegraphics[width=0.9\textwidth]{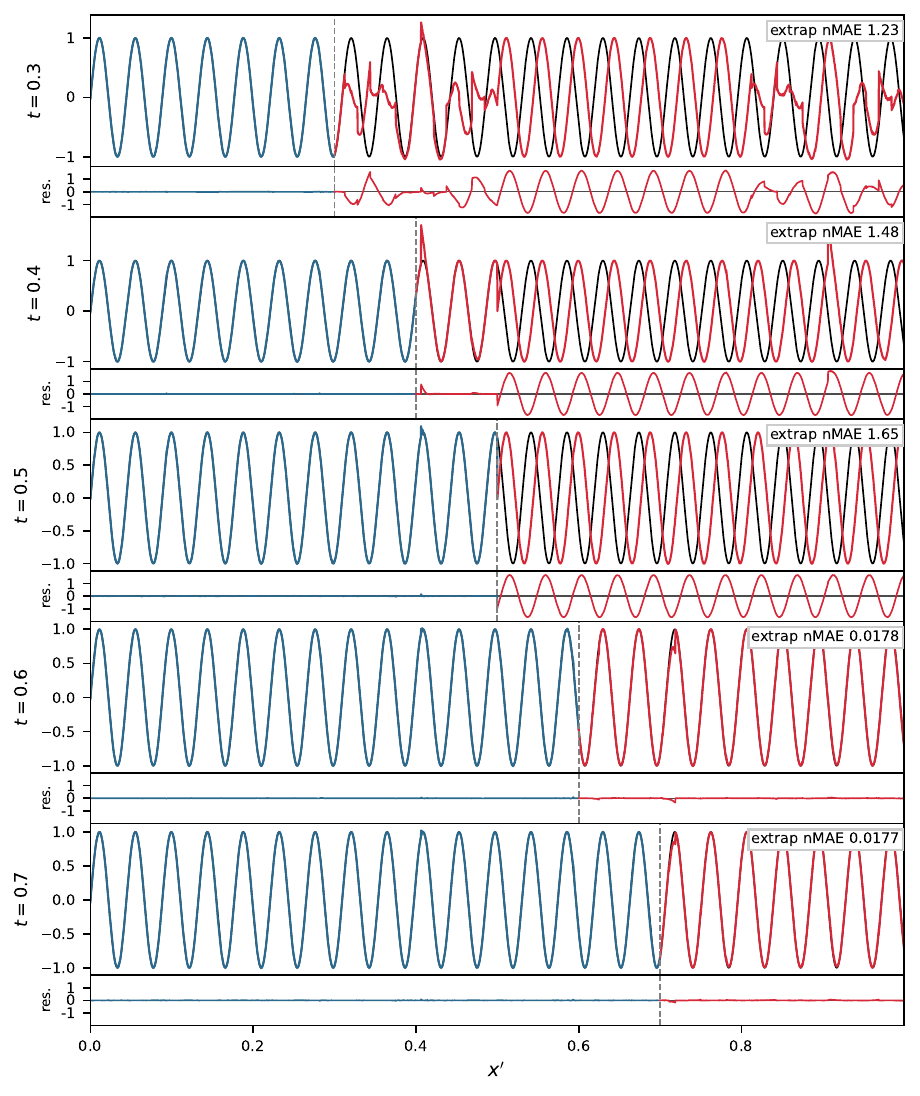}
\caption{For $X'_\text{train} < t < X'_\text{extrap}$, from top to bottom, predictions and residuals for the training data (blue) and the extrapolation region (red) for $t=0.3$, $t=0.4$, $t=0.5$, $t=0.6$, and $t=0.7$.}
\label{fig::vsplit}
\end{figure}

Investigating the limits imposed by Equation \ref{eqn::nb2e_ineq}, we examined the sensitivity of NB2E in our examples from Section \ref{sec::examples} to the maximum normalized value of the training domain.  We split the normalized data set $X'$ into $X'_\text{train}$ and $X'_\text{extrap}$. The total number of samples (10{,}000 on [0, 1)) and the cycle count $c=2^{4.5}$ are held fixed, so the number of training samples and the number of cycles in $X'_\text{train}$ both scale with $t$.  For $X'_\text{train} < t < X'_\text{extrap}$, we tested examples at increments of 0.1 from $t=0.3$ through $t=0.7$, training each for 1,500 epochs. The results for the sine function are shown in Figure \ref{fig::vsplit}, with $t$ identified in each plot by the gray dashed vertical line.  Predictions for $X'_\text{train}$ are in blue and predictions for $X'_\text{extrap}$ are in red.  All sensitivity experiments use the sine at $c=2^{4.5}$, a non integer cycle count, so that they characterize the general case rather than the special case of an integer.

It can be easily seen that the extrapolation fails for any $t\leq0.5$, as suggested by Equation \ref{eqn::nb2e_ineq}.  As we discussed in Section \ref{sec::limit}, valid prediction cannot be made unless the MLP has been trained with each element of the NB2E vector.  Training with $t <0.5$ does not train the first element of the NB2E vector.  So, as expected for examples $t=0.5$, $t=0.4$, and $t=0.3$, the prediction fails at or prior to $x'=0.5$.  At $t=0.6$ and $t=0.7$, as with our initial example in Section \ref{sec::examples}, the model has clearly learned the signal and extrapolates nicely.  So, we can see rather clearly that the conditions of Equation \ref{eqn::nb2e_ineq} are critical to using NB2E and, in practical application, $t$ should be as high as possible to ensure proper training of all the bit interrelationships of the NB2E while still leaving space for validation and extrapolation.

\begin{figure}[h]
\centering
\includegraphics[width=0.9\textwidth]{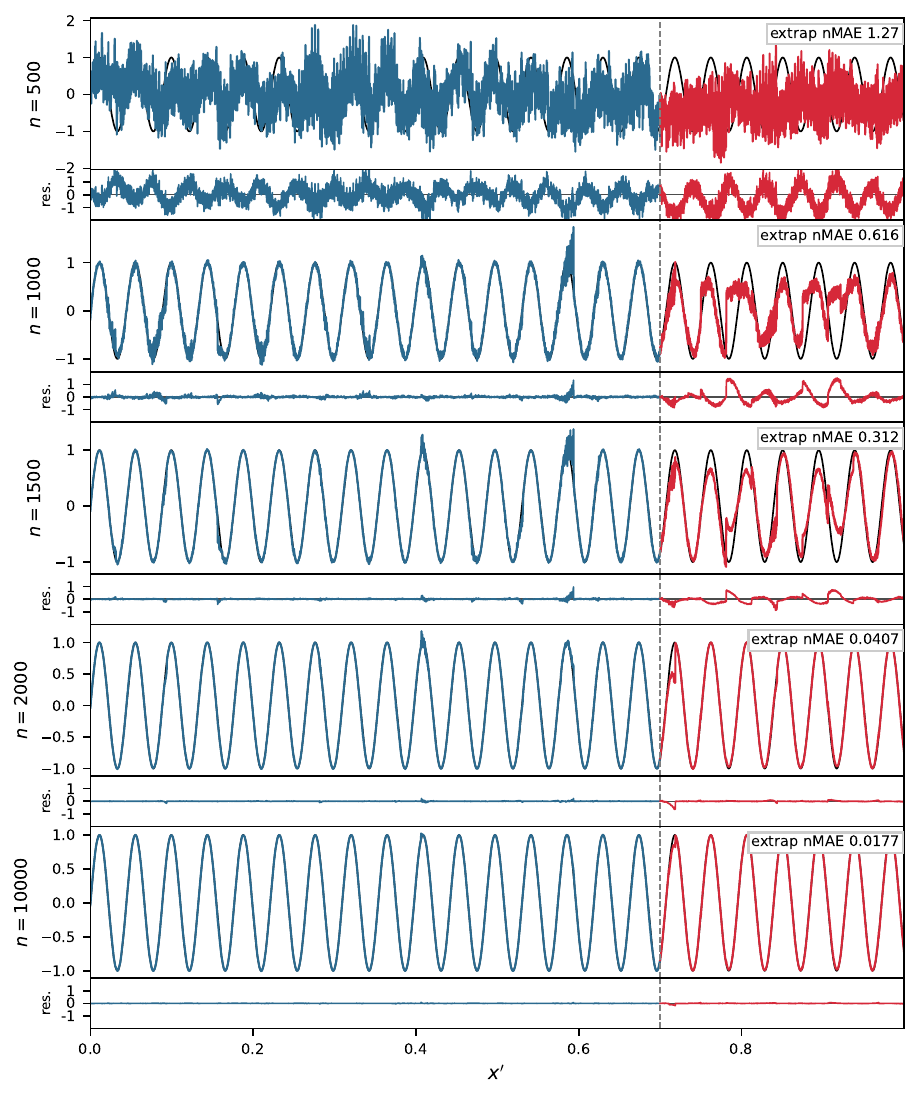}
\caption{({\em upper panels}) Predictions of the sine curve at $c=2^{4.5}$ for the training data (blue) and the extrapolation region (red) shown with the true function (black).  ({\em lower panels}) Residuals of the predictions. The number of data points in each data set are labeled to the left of each panel. The dashed gray vertical line is shown at $t=0.7$.}
\label{fig::sensitivity}
\end{figure}

We also investigated the sensitivity of the NB2E method to the number of data points. In our examples from Section \ref{sec::results}, we drew 10{,}000 samples in each case. The results of our analysis are shown in Figure \ref{fig::sensitivity}, in which we provide panels showing the training and extrapolation predictions with 500, 1{,}000, 1{,}500, 2{,}000, and 10{,}000 samples. At $c = 2^{4.5}$ the requirement is substantial. With 500 samples the network does not fit the training data, and with 1{,}000 and 1{,}500 samples the extrapolation error is 0.62 and 0.31. From 2{,}000 samples, which is roughly 1{,}400 in the training region or 90 per training cycle, the predictions improve smoothly and reach 0.018 at 10{,}000.

\begin{figure}[h]
\centering
\includegraphics[width=0.9\textwidth]{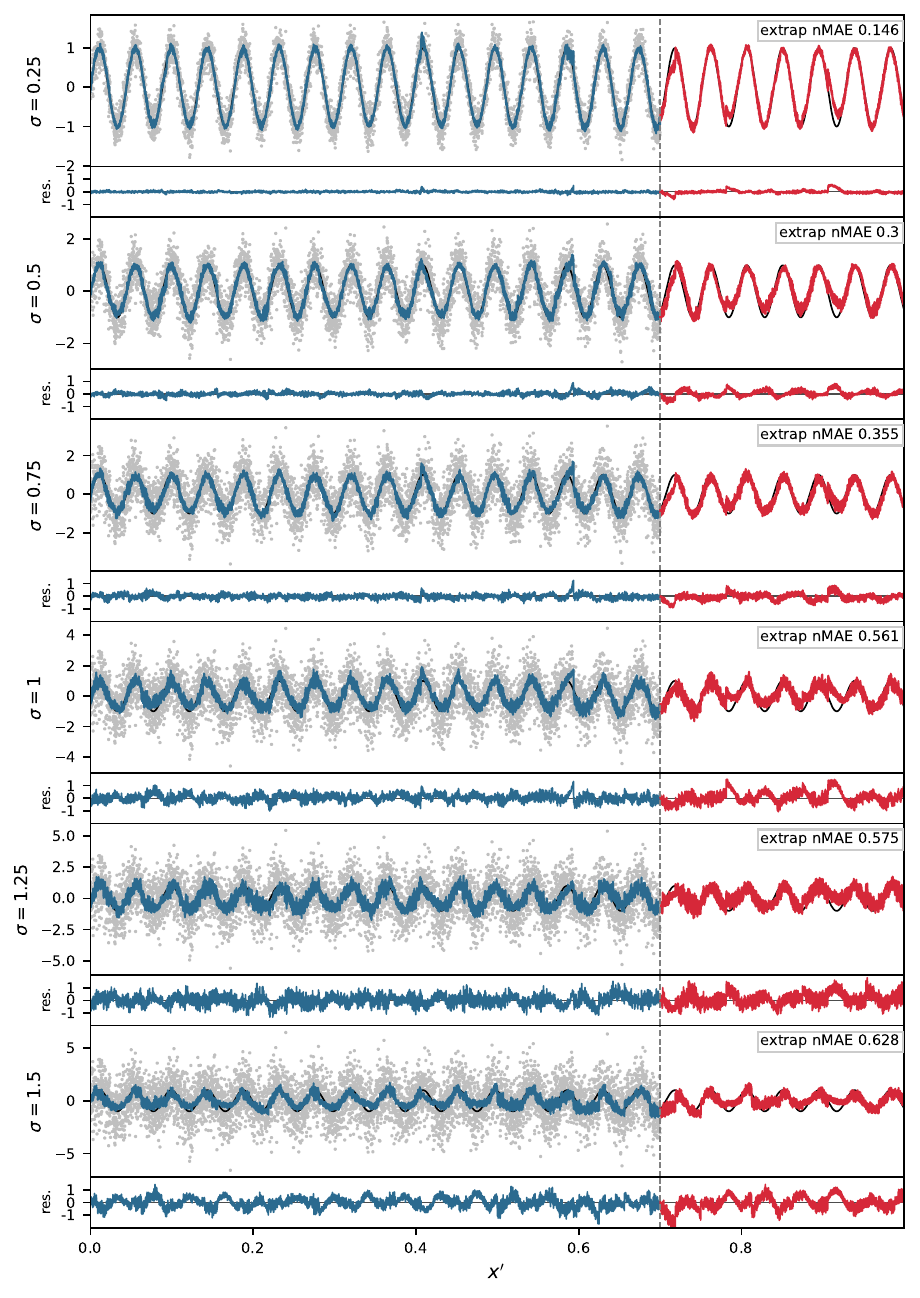}
\caption{({\em upper panels}) Predictions of the sine curve at $c=2^{4.5}$ for the training data (blue) and the extrapolation region (red) shown with the true function (black).  ({\em lower panels}) Residuals of the predictions.  Standard deviation of the Gaussian noise is shown to the left of each panel.}
\label{fig::noise}
\end{figure}

Considering applicability of the NB2E method to noisy real-world signals, we also examined the ability of the MLP to extrapolate a sine curve with added random noise. We applied Gaussian noise with zero mean and standard deviation in the range 0.25 to 1.5 in increments of 0.25. The results are shown in Figure \ref{fig::noise}. The extrapolation retains the period and phase of the sine at every noise level tested, with the amplitude and shape degrading as $\sigma$ grows, from nMAE of 0.15 at $\sigma = 0.25$ to 0.63 at $\sigma = 1.5$. Residuals and nMAE in both regions are computed against the clean sine, and the noisy training samples are shown in gray.

\clearpage
\section{Discussion}
\label{sec::discussion}

NB2E demonstrates the ability to successfully extrapolate periodic functions beyond their training bounds, with some difficulty when $c$ is at or near odd integers. This capability is distinct from existing approaches: symbolic regression \citep{schmidt2009distilling,martius2016extrapolation,udrescu2020ai} discovers explicit functional forms through search, Physics-informed Neural Networks \citep[PINNs,][]{raissi2019physics} leverage known governing equations, and Neural ODEs \citep{chen2018neural} model temporal derivatives but do not inherently capture periodicity. NB2E enables extrapolation of sine, triangle, sawtooth, and square periodic functions through input encoding alone, without architectural modifications or prior knowledge of the functional form.

Where NB2E succeeds, its residual does not grow with distance from $t$ (Figures \ref{fig::sine24} and \ref{fig::vsplit}), whereas the FFE prediction at $c=2^{4.5}$ in Figure \ref{fig::ex} drifts increasingly from the target across the region and sequence-based models tend to accumulate error with each additional step in the prediction horizon.  Where NB2E fails, at $c=17$ in Figure \ref{fig::ex}, its residual does grow.  We report this as an observation on these examples and do not generalize it. 

Expanding on this concept, another valuable characteristic of the NB2E MLP predictions is the continuity thereof.  Sequence-based methods make predictions at fixed intervals, whereas predictions with NB2E can be made for the entire set of representable numbers less than unity.  This feature, which any coordinate encoding shares, lends itself to use with {\em irregular} time series data.  Sequence prediction models tend to ignore the actual value of the progression axis (generally time or position) by assuming regular sampling.  In this manner, these models learn dependencies on {\em order} rather than any relationship with the axis.  This limits any prediction to the regular sampling cadence.\footnote{In applications where regularity is inherent, a regular cadence is of course a feature rather than a limitation.}

The contrast with Fourier encoding is instructive. In our tests, both NB2E and FFE provide multi-scale decompositions, yet NB2E performs near-universally better than FFE with the exceptions of even integers, where FFE is slightly better, and narrow bands near the odd integers adjacent to powers of two, where NB2E has its largest errors and FFE partially recovers.  Both the waveform at each index and the set of values differ between the two encodings (see Figure \ref{fig::freq}), and these experiments do not separate their effects. We leave that question, and the mechanism, to future work.  

We state the limits of the scope of our effort plainly. All targets are synthetic and exactly periodic, and each network is trained on exactly one signal. We have not tested real, approximately periodic data, and we regard that as the next step rather than part of this study. We have also not tested multidimensional signals or encodings, and we make no suggestions regarding whether or not such an extension would be successful. The multi component signals of an earlier version of this work were not retested under the revised protocol and are outside the present scope.

The prediction limits that we described in Section \ref{sec::limit} as well as additional limitations found in the sensitivity analysis in Section \ref{sec::sensitivity} are substantial constraints on the use of NB2E.  Due to the necessity of learning the relationships between each element of the encoding, a neural network trained with NB2E input can only predict forward to $x' = 1$, that is to $x = z$ in the units of the data.  In Section \ref{sec::limit} we discussed a method of normalization to render this limitation much less obstructive.  We also suggest that future work could establish methods of working with a base-2 input vector in terms of relative rather than absolute positions.  That is, to learn relationships between relative positional differences \citep[as in][for transformer positional encoding]{shaw2018selfattention} in the vector such that powers of two do not form boundaries in the prediction space.  

\section{Conclusions}
\label{sec::conclusions}

We have demonstrated that NB2E enables neural networks to extrapolate periodic sine, triangle, sawtooth, and square functions beyond their training bounds. Using NB2E, vanilla MLPs successfully extrapolate four periodic waveforms, performing better than FFE except at even integer cycle counts, where FFE is slightly better, and near odd integer cycle counts adjacent to powers of two, where NB2E has its largest errors, while continuous inputs fail at this task despite identical network architecture and training procedure.

Through empirical characterization, we established the requirements and limitations of this approach. Extrapolation requires the training domain to extend beyond 0.5 in normalized space, is most accurate when the number of cycles over the normalized domain is an even integer, and remains valid to a limit of unity. While this seems rather constraining, we identified in Section \ref{sec::limit} that normalization need not be to the maximum value of the dataset, leaving ample room for predictive extrapolation. For the sine function at the cycle count tested, the method degrades gracefully with additive noise and requires on the order of 2{,}000 samples, and in the examples where it succeeds, its residual does not grow with distance from the training boundary, an observation we do not generalize beyond these examples.

While we have characterized when and how this capability emerges empirically, the theoretical question of why discrete coordinate encodings fundamentally change the learning dynamics remains open. Understanding the mathematical principles and formalizing the conditions under which binary representations enable extrapolation are important directions for future work. Additionally, extensions to multi-dimensional inputs and integration with modern architectures beyond vanilla MLPs should be examined. 

\section*{Acknowledgments}
The authors acknowledge the NASA Goddard Space Flight Center Sciences and Exploration Directorate for providing funding for this research.  

Resources supporting this work were provided by the NASA High-End Computing (HEC) Program through the NASA Center for Climate Simulation (NCCS) at Goddard Space Flight Center.

\clearpage
\bibliography{refs}
\bibliographystyle{authoryear}

\end{document}